\documentclass[11pt]{article}
\usepackage{geometry}
\usepackage{coling2020}
\usepackage{times}
\usepackage{url}
\usepackage{latexsym}
\usepackage{microtype}
\usepackage{graphicx}
\usepackage{subfig}
\usepackage{booktabs}
\usepackage{comment}
\usepackage{multirow,tabularx}
\usepackage{hyperref}
\colingfinalcopy 

\title{DSC IIT-ISM at SemEval-2020 Task 8: Bi-Fusion Techniques for Deep Meme Emotion Analysis}

\author{
\\
  \\
  \And
  Pradyumna Gupta*, Himanshu Gupta*, and Aman Sinha \\
  \\
  Indian Institute of Technology (Indian School of Mines) Dhanbad, India \\
  {\tt \{pradyumna.gupta22, hg24091996, amansinha091\}@gmail.com} \\
  \And
  \\
  \\
  \\}

\begin{document}
\maketitle
\begin{abstract}
  Memes have become an ubiquitous social media entity and the processing and analysis of such multimodal data is currently an active area of research. This paper presents our work on the Memotion Analysis shared task of SemEval 2020, which involves the sentiment and humor analysis of memes. We propose a system which uses different bimodal fusion techniques to leverage the inter-modal dependency for sentiment and humor classification tasks. Out of all our experiments, the best system improved the baseline with macro F1 scores of 0.357 on Sentiment Classification (Task A), 0.510 on Humor Classification (Task B) and 0.312 on Scales of Semantic Classes (Task C).
\end{abstract}

\section{Introduction}
\label{intro}

Social media is becoming increasingly abundant in multimedia content with internet memes being one of the major types of such content circulated online. Internet memes, or simply, memes are mostly derived from news, popular TV shows and movies making them capable of conveying ideas that the masses can understand readily. Today memes are shared in numerous online communities and are not only limited to one language. It is therefore imperative to study this emerging form of  mass communication. Although they are mostly for amusement and satire, memes are also being used to propagate controversial, hateful and offensive ideologies. Since it is not feasible to manually inspect such content it is important to build systems that can process memes and segregate them into appropriate categories.

The SemEval 2020 shared task on Memotion Analysis \cite{chhavi2020memotion} aims to bring attention to the study of sentiment and humor conveyed through memes. The challenge involves 3 tasks:
\begin{itemize}
    \item Task A involves classifying the sentiment of an internet meme as positive, negative or neutral.
    \item Task B is the multilabel classification of a meme into 4 categories viz. sarcastic, humorous, offensive and motivational.
    \item Task C is to quantify, on a scale of 0-3 (0 = not, 1 = slightly, 2 = mildly, 3 = very), the extent of sarcasm, humor and offence expressed in a meme.
\end{itemize}
Here, we consider Task B as a set of 4 independent binary classification subtasks while Task C comprises 3 subtasks each being a multiclass classification problem.

This paper describes our work on all three tasks. Here, a meme has two modalities - textual and visual cues. We use a combination of state-of-the-art model architectures and adapt them for processing multimodal memes through transfer learning while also training models from scratch for comparison. We explore various modality fusion techniques and also the inter-task dependency. Our systems$^{\dagger}$ score higher than the baseline results released on all the 3 tasks. We investigate our model performances from relative and absolute perspectives and also explore some possible improvements. We make our code publicly available.

%
%
\blfootnote{
    %
    %
    %
    %
    %
    %
    \hspace{-0.65cm}  
    \textit{*} Equal contribution.\\
    {$\dagger$} Submissions have been made under the username of \textit{hg} on CodaLab.\\
    This work is licensed under a Creative Commons 
    Attribution 4.0 International License.
    License details:
    \url{http://creativecommons.org/licenses/by/4.0/}.
}

\section{Related Work}
\label{related}

Discussing the sentiments of memes has been a relatively new research area. Although interest in the research related to meme generation, circulation and perception has been increasing in the last couple of years where studies of \newcite{10.1145/3371158.3371403}, \newcite{wang-wen-2015-cheezburger} have shown pioneering work, studies exclusively analysing meme emotion are relatively scarce.

With the explosion of social media in recent years and human generated content on them like tweets, memes etc., deep learning techniques have captured the attention of researchers due to their ability to significantly outperform traditional methods in sentiment analysis tasks. Studies based on discussing textual segments \cite{DBLP:journals/corr/abs-1801-07883} and based on other channels like images and videos \cite{poria-etal-2017-context} have been in progress. Visual sentiment and emotion analysis, although having received less attention than text based analysis, has involved studies of \newcite{10.1145/2502081.2502282}, \newcite{10.5555/3298239.3298274} and \newcite{10.1145/3368567.3368582} where features extracted from images using deep learning have been used for predictions.

Apart from the unimodal methods we have seen so far, the multimodal methods involving simultaneous processing of textual and visual information are of special interest, as studies by \newcite{5540120} and \newcite{DBLP:journals/corr/ZahavyMKM16} have reported improvement  in performance by combining modalities. Multimodal processing of features using hierarchical methods \cite{DBLP:journals/corr/abs-1806-06228,DBLP:journals/corr/abs-1810-03414} and attention based methods \cite{Huang2019MultiHeadAW,moon-etal-2018-multimodal-named} have shown to be effective fusion techniques and have been used by teams in previous iterations of SemEval as well. Multitask learning for processing multimodal information has also been studied by works like \newcite{DBLP:journals/corr/abs-1905-05812}, where the authors have tried to capture inter-dependence between multimodal tasks like emotion recognition and sentiment analysis. In addition to above, processing of multimodal information specifically for memes have also been studied by some papers which have greatly inspired our work, including papers by \newcite{French2017ImagebasedMA}, which discusses correlation between the implied semantic meaning of image-based memes and the textual content of discussions in social media and also \newcite{10.1145/3343031.3350939}, which have developed models to generate rich face emotion captions from memes.

\section{Data}
\label{data}

The dataset \cite{chhavi2020memotion} consists of almost 7K meme images along with their extracted text. Each sample has a sentiment annotation into one of 3 classes- negative (0), neutral (1) or positive (2). There are another 3 semantic classes namely humor, sarcasm and offence with their fine grained labels as not (0), slightly (1), mildly (2) and very (3). Also there is a binary column denoting whether a meme is motivational. Another set of around 2K samples form the test set. We split the dataset into train and validation sets of around 6K and 1K samples respectively. From the data distribution in Table \ref{tab:data}, we can see that all the labels in the dataset have considerable class imbalance.

\begin{table}[h!]
  \begin{center}
    \begin{tabular}{lcccc|cccc}
      \toprule 
      \multirow{2}{*}[-2.5pt]{\bf{Label}} & \multicolumn{4}{c|}{\bf{Train Set}} & \multicolumn{4}{c}{\bf{Validation Set}}\\
      \cmidrule{2-9}
    & \bf{0} & \bf{1} & \bf{2} & \bf{3} & \bf{0} & \bf{1} & \bf{2} & \bf{3}\\
      \midrule 
Sentiment & 519 & 1894 & 3530 & - & 112 & 307 & 630 & - \\
Humor & 1400 & 2060 & 1952 & 531 & 251 & 392 & 286 & 120 \\
Sarcasm & 1304 & 2983 & 1323 & 333 & 240 & 524 & 224 & 61 \\
Offence & 2289 & 2203 & 1259 & 192 & 424 & 389 & 207 & 29 \\
Motivation & 3844 & 2099 & - & - & 681 & 368 & - & - \\
      \bottomrule 
    \end{tabular}
  \end{center}
  \caption{Dataset label distributions for the train and validation sets.}
  \label{tab:data}
\end{table}

\section{Methodology}
\label{method}

A meme typically consists of an image paired with a catchphrase on it. The image is generally derived from some popular source and is reused as a template to represent an underlying theme. The text on the other hand is usually original and specific, written by the creator, and it is what makes the meme unique. Together they convey some sentiment  and emotion which we study here. In this task we focus on three main approaches:

\begin{itemize}
    \item Textual, based on the text part of the meme.
    \item Visual, based on the image of the meme.
    \item Combined or fused, unifying the textual and visual inputs in a single end-to-end model.
\end{itemize}
A more detailed explanation of our approaches can be found in the following subsections.

\subsection{Textual Features}
Similar to textual content on social media from other sources like tweets and comments, the textual part of a meme also contains important cues that can help extract the sentiment and emotion of the meme. Since the textual part of a meme gives it some specific context, it can sometimes be more detrimental in analysing the sentiment or emotion behind a meme. Therefore, as the first approach in our work we analyse some of the text based models to solve the three tasks.

{\bf BiLSTM}: Long Short-Term Memory (LSTM), \cite{10.1162/neco.1997.9.8.1735} is a widely known recurrent neural network (RNN) architecture. We use Bidirectional LSTM \cite{Schuster1997BidirectionalRN} models for our experiments. A Bidirectional LSTM (BiLSTM) layer processes the text in the forward as well as backward direction and hence is known to give better context understanding.


{\bf RoBERTa}: Bidirectional Encoder Representations from Transformers, or BERT \cite{DBLP:journals/corr/abs-1810-04805} has been a revolutionary language model for a while now which is based on the self-supervised pretraining technique and has shown to generalize extremely well to downstream tasks. The recently released RoBERTa - A Robustly Optimized BERT Pretraining Approach \cite{DBLP:journals/corr/abs-1907-11692}, builds on BERT’s language masking strategy and modifies key hyperparameters in BERT, including removing BERT’s next-sentence pretraining objective, and training with much larger mini-batches and learning rates. RoBERTa has also been trained on a much larger text corpus than BERT, for a longer amount of time. This allows RoBERTa representations to generalize even better to downstream tasks compared to BERT. Here we use Roberta-Base to encode the textual part of the meme.

\subsection{Visual Features}
It is well known that the human brain prefers and is faster at processing visuals  as compared to reading text. This is mainly because the entire image is perceived at once while text has to be processed linearly. On seeing a meme, the image is the part one perceives first and so it can provide a generic context on top of which the text part acts. We extract the visual representation of images in the form of abstract as well as specific features.

{\bf AlexNet}: AlexNet \cite{NIPS2012_4824} is a widely known convolutional neural network (CNN) which consists of convolutional and fully connected layers and is one of the first networks to use rectified linear unit as its activation function. We do not use a pretrained model and train from scratch on the dataset.

{\bf ResNet}: ResNet \cite{DBLP:journals/corr/HeZRS15}, short for Residual Networks is a classic neural network used as a backbone for many computer vision tasks. ResNet makes use of the skip connection to add the output from an earlier layer to a later layer thereby helping it mitigate the vanishing gradient problem. For the sake of experiments reported in this paper we use ResNet with its ImageNet \cite{Deng2009ImageNetAL} pretrained weights. 

{\bf Facial Expressions}: On deeper exploratory analysis of the given dataset we found that most of the meme images contain faces of one or more people in them. The facial expression of these people might contain useful information about the overall sentiments expressed by the meme. \newcite{10.1145/3343031.3350939} have already implemented an intuitive end-to-end model for generating rich face emotion captions from meme images using a ResNet to generate emotion features which are used by a captioning model to generate caption sequences. Therefore, as a part of our visual approach, for capturing facial expressions in our dataset we try using their pretrained ResNet block for generating predictions of the current task. However, because of poor performance of  this model on the validation set we exclude it from our experiments on the test set.

{\bf Face Emotion Captions}: Unlike our earlier approach of processing facial emotions in meme images, this time we try incorporating facial expressions as a third channel of information in addition to the image and meme overlay text. This is done by building our textual models on top of the text captions generated by the \newcite{10.1145/3343031.3350939} model and then treating these textual models as a third input in our fusion approaches, described later. It is also worthwhile to mention here that for the memes that did not contain faces, random captions were generated by the model. However again due to poor performance of these models on the validation set we exclude this from our experiments on the test set.

\subsection{Multimodal Approach}
\begin{figure}[htp]
    \centering
    \subfloat[]{{\includegraphics[scale=0.42]{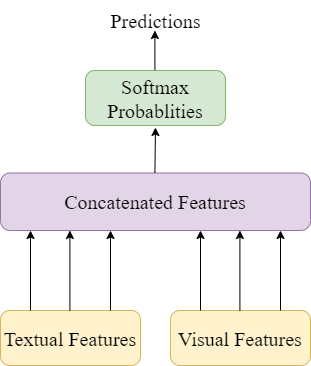} }}
    \qquad
    \subfloat[]{{\includegraphics[scale=0.42]{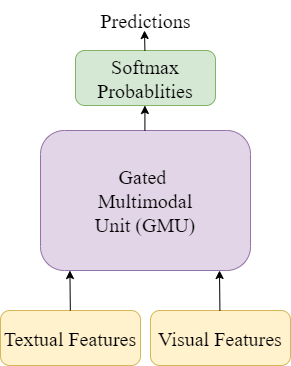} }}
    \qquad
    \subfloat[]{{\includegraphics[scale=0.42]{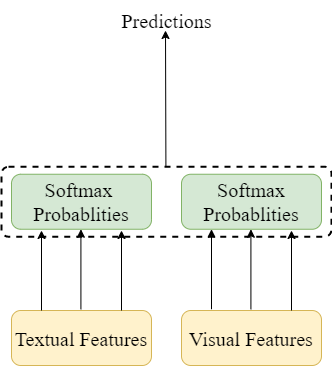} }}
    \caption{Modality fusion strategies- (a) Early Fusion, (b) GMU and (c) Late Fusion.}
    \label{fig:fusion}
\end{figure}

The above methods consider both modalities individually for representation learning. However, effective computational processing of internet memes needs a combined approach. After extracting the text and image features they need to be combined to get a complete representation of a meme. Figure \ref{fig:fusion} gives a high level illustration of the modality fusion techniques that we use.

{\bf Early Fusion}: This method involves simply concatenating the textual and visual feature vectors extracted by above mentioned models to produce a joint representation before passing it through a softmax classifier. We do this for the features generated by BiLSTM \& AlexNet, and also for RoBERTa \& ResNet.

{\bf GMU}: The Gated Multimodal Unit \cite{Ovalle2017GatedMU} is a building block to be used as an internal unit in a neural network architecture whose purpose is to find an intermediate representation based on a combination of data from different modalities. The GMU learns to decide how modalities influence the activation of the unit using multiplicative gates. We use GMU particularly to combine the representations produced by RoBERTa and ResNet.

{\bf Late Fusion}: In this method, instead of combining text and visual models at the time of training, we fuse the models at inference time by using their individual softmax prediction scores. During our experiments we assign equal weights to text and visual models to take effectively an average of the prediction scores of the models to make the final predictions. We do this to combine the predictions of RoBERTa with ResNet.

\subsection{Multitask Learning}
In all the above approaches we have to train separate models for each subtask of the three tasks thereby requiring a total of 8 models to be trained per submission. Using multitask learning (MTL) enables us to train a model with multiple outputs which can be used for multiple classification tasks simultaneously. In MTL we have the hidden layers of the network shared among the tasks and also keep some individual layers for each output. MTL is known to leverage useful information contained in multiple related tasks to help improve the generalization performance of all the tasks \cite{DBLP:journals/corr/ZhangY17aa}. To this end, we use multitask models with early fusion of features from BiLSTM \& AlexNet for Task A and the 4 subtasks of Task B together. We also use a similar model for solving Task A and the subtasks of Task C together.

\section{Experiments}
\label{exp}

Using the techniques mentioned in the previous section we train several models for our submissions. In this section, we provide the training parameters for reproducibility\footnote{Link to our code- \url{https://github.com/dsciitism/SemEval-2020-Task-8}} and also their results on the test data.

\subsection{Training Details}
Except for the MTL model, we train the following models separately for Task A and each of the subtasks of Task B \& Task C. We use Keras with Tensorflow backend \cite{chollet2015keras} on Google Colab GPU to implement and train our models. Unless otherwise stated, all models use the default Adam optimizer with the Cross Entropy loss and trained till the validation loss stopped improving. We use a batch size of 32.

{\bf BiLSTM}: As a preprocessing step, we remove the punctuation and symbols from the text data, convert to lowercase and tokenize to generate sequences with truncated length of 64. We train models with an input layer having embedding size 32, followed by a bidirectional LSTM layer with recurrent dropout and then a softmax dense layer. We keep the number of LSTM units to be 32 as higher values didn't improve performance. Also, we apply a dropout of 0.5 in between the layers. To deal with the class imbalance, we use appropriate class weights for loss during training.

{\bf RoBERTa}: For implementation details of RoBERTa-Base readers can refer to \newcite{DBLP:journals/corr/abs-1907-11692}. Since transformer based models can handle long sequences we take the input sequence length for RoBERTa equal to the length of the longest text which was 199. Apart from preprocessing steps like lower casing, punctuation and symbol removal, we use oversampling to make the target class ratio 1 for handling class imbalance as the model tends to get biased towards one class on the given dataset. We fine tune the pretrained model on the given dataset using the output corresponding to the CLS token of the last layer followed by a batch normalization layer, a dense layer of hidden size 256 and finally a softmax layer. The model is trained with a learning rate of 3e-4.

{\bf AlexNet}: As mentioned in the original paper we resize the images to 224 x 224 with 3 channels as input to the model. The filter sizes of the convolutional layers follow [96, 128, 128, 256, 256]. After flattening we apply 2 dense layers of size 1024 each before the final softmax layer. We apply a dropout of 0.4 after each layer and use class weights to handle imbalance.

{\bf ResNet}: Before passing the meme images to the model all the images are resized to 256 x 256, rescaled and all the three RGB channels of the images are retained. We use the pretrained ImageNet weights to initialize the model and the outputs are followed by dense layers of hidden sizes [768, 256] and a softmax layer. A learning rate of 3e-4 is used to train the model.

{\bf BiLSTM+AlexNet}: After using the same process and architecture as mentioned in their individual training details, we replace the softmax layer in the BiLSTM and the AlexNet model with a dense layer of 64 units to get the text and image vector representations. Then we pass both through a shared dense layer of size 64 and concatenate (Early Fusion) the outputs to form a joint vector of length 128. Then we keep another dense layer of 128 units before the softmax prediction layer. Here also we employ class weight balancing.

{\bf RoBERTa+ResNet}: For the pretrained models we have explored three ways of fusion of information, as mentioned earlier, viz Early Fusion, GMU and Late Fusion. For Early Fusion (Figure \ref{fig:fusion}(a)), we batch normalize the outputs of the models, concatenate them and finally pass them through a softmax layer. For GMU (Figure \ref{fig:fusion}(b)), we take a bimodal GMU unit which is connected to the two models as input and has a softmax layer connected at its output. Finally, for Late Fusion (Figure \ref{fig:fusion}(c)), we take the predictions of the two earlier separately trained models and average their prediction probabilities to get the final predictions. Since both the pretrained architectures are relatively deeper neural networks we use SGD optimizer while training the Early Fusion and GMU models to prevent gradient vanishing and oversampling to prevent the model from getting biased.

{\bf MTL}: For the multitask learning approach we use the BiLSTM+AlexNet architecture, with slight modifications, and train from scratch. We train 2 models here, one for Task A with Task B and another for Task A with Task C. To handle the increased complexity of the model objective we add another bidirectional LSTM layer and double the vector size for each modality to 128. We also increase the size of the joint vector to 256 which is fed into separate 128 unit dense layers for each model task followed by individual softmax layers. For training the model we keep the loss contribution or weights equal for all outputs. We also try using the Task B semantic labels for auxiliary tasks with Task A as the main objective but it doesn't give any improvements.

\begin{table}[h!]
  \begin{center}
    \begin{tabular}{lcc|ccc}
      \toprule 
      \textbf{Model} & \textbf{Modality} & \textbf{Fusion Strategy} &\textbf{Task A} &\textbf{Task B} &\textbf{Task C}\\
        \midrule 
        Baseline Results & - & - & 0.218 & 0.500 & 0.301 \\
        \midrule 
        BiLSTM & Text & - & 0.319 & 0.502 & 0.290 \\
        RoBERTa & Text & - & 0.314 & 0.494 & 0.276 \\
        AlexNet & Image & - & 0.309 & 0.490 & 0.302 \\
        ResNet & Image & - & 0.324 & 0.508 & {\bf0.312} \\
        \midrule 
        BiLSTM+AlexNet & Text \& Image & Early Fusion & 0.323 & 0.495 & 0.291 \\
        BiLSTM+AlexNet (MTL) & Text \& Image & Early Fusion & 0.322 & 0.495 & 0.267 \\
        \midrule 
        RoBERTa+ResNet & Text \& Image & Early Fusion & {\bf0.357} & {\bf0.510} & 0.306 \\
        RoBERTa+ResNet & Text \& Image & GMU & 0.346 & 0.503 & 0.303 \\
        RoBERTa+ResNet & Text \& Image & Late Fusion & 0.321 & 0.508 & {\bf0.312} \\
      \bottomrule 
    \end{tabular}
  \end{center}
  \caption{\label{tab:results} Model details and their macro F1 scores on the test set. For Task B and Task C, the macro F1 scores for each subtask were averaged.}
\end{table}

\subsection{Results}
The evaluation metric used here is the macro F1 score for Task A. For Task B and Task C, the macro F1 scores for each subtask were averaged. Table \ref{tab:results} has the model test set scores from our experiments as well as the baseline results released by the competition organisers. From the table we can see that the RoBERTa+ResNet model with Early Fusion has the best performance for Task A and Task B while the same model using the Late Fusion strategy has the best score for Task C. For the BiLSTM+AlexNet model, using MTL did not give any improvements. We find that among the 3 fusion strategies, Early Fusion scores best for Task A and Task B, Late Fusion gives the best performance on Task C while GMU gives intermediate results. We chose the BiLSTM+AlexNet model for our contest submission as it gave the best results during the development phase. However, after final evaluation the RoBERTa+ResNet models were found to be better.

\section{Discussion}
\label{analysis}

In order to justify the results and to discuss some observations we have about the dataset and memes in general, we present a detailed analysis in this section. We investigate why the state-of-the-art pretrained models could not give the superior results that is expected from them and highlight the areas of potential improvement.

\begin{figure}
    \centering
    \subfloat[]{{\includegraphics[scale=0.27]{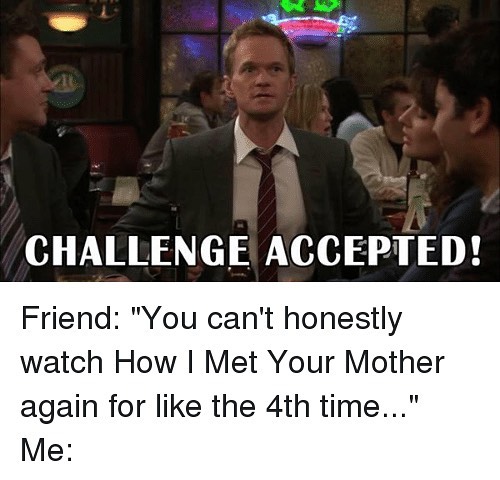} }}
    \qquad
    \subfloat[]{{\includegraphics[scale=0.5]{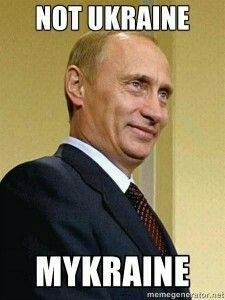} }}
    \qquad
    \subfloat[]{{\includegraphics[scale=0.22]{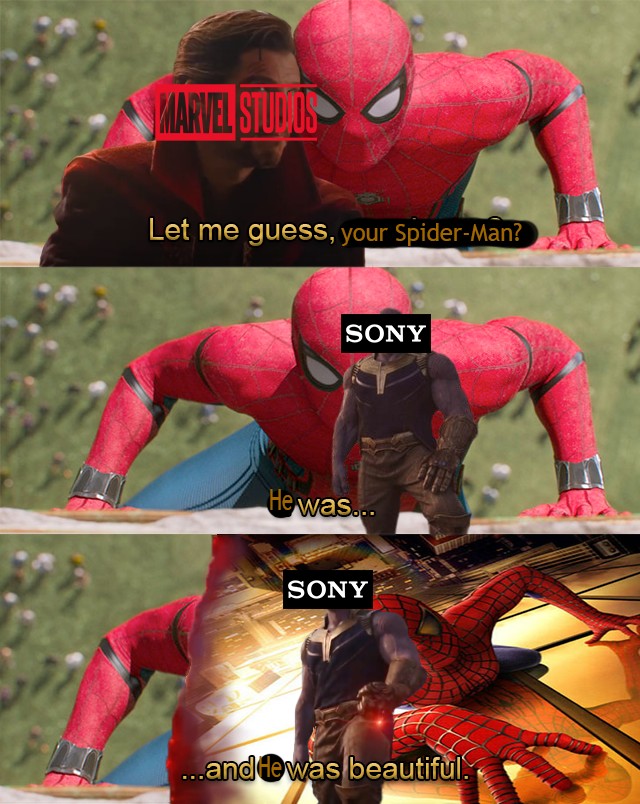} }}
    \caption{Some examples from the dataset that highlight the unconventional aspects of memes.}
    \label{fig:memes}
\end{figure}

Apart from deep learning models we also try the TF-IDF model with classical machine learning classifiers like logistic regression. Although the deep learning models perform better, the validation set score difference is not large. On examining the data we find examples in which the text is scattered around the image and since an OCR is used for preparing the dataset, it ended up giving jumbled phrases in an order which is different from what is intended by the maker of the meme. An example of this can be seen in the Figure \ref{fig:memes}(a) where the phrase \textit{"CHALLENGE ACCEPTED!"} will appear at the beginning of the text when read by an OCR whereas in reality it is intended to be perceived at the end of the dialogue when read by a human. Such samples undermine the advantage that sequential models have over frequency based models.

For the unimodal BiLSTM model, apart from training the embedding layer parameters from scratch we also use the pretrained GloVe embeddings \cite{pennington2014glove}. Although the GloVe embeddings are trained on a much larger corpus than the dataset here, we found that it did not improve the performance of the text model. This can be attributed to the difference in the vocabulary usage of memes over standard English and the intentional misspellings used for common words, as seen in Figure \ref{fig:memes}(b) which uses the word \textit{“MYKRAINE”} to imply \textit{“My Ukraine”}. Moreover, the incorrect grammar used in a lot of memes may also be the reason why RoBERTa, despite being pretrained on a huge English corpus, failed to generalize well over the meme language and scored below BiLSTM for all the three tasks. Using character level embedding approaches to the language model can be helpful in dealing with such cases.

For the image models also, in spite of having an edge over AlexNet, the pretrained deep models, Inception \cite{DBLP:journals/corr/SzegedyLJSRAEVR14} and ResNet, which we try, didn't outperform it by a large margin. We believe this is because there is very little similarity between the ImageNet dataset and the meme images used here. Memes are usually images that have been edited and also have some text overlay which can act as noise for the network thereby hindering representation learning. An example of this is Figure \ref{fig:memes}(c) which has heavy image editing and a lot of text. To deal with the text overlay in memes, some techniques in image masking can be used as a preprocessing step to replace the text with its background.

Overall, the winning macro-averaged F1 scores in the competition for Task A, Task B and Task C were less than 0.36, 0.52 and 0.33 respectively which shows there is ample room for improvement. Out of our three fusion strategies, simple concatenation (Early Fusion) turned out to be the best. But when we see a meme both text and image rely on each other to jointly convey an idea. We strongly feel that there is a need for a better fusion strategy or learning technique that can generate a unified and coherent representation, which is analogous to how a human brain processes memes. For the image captioning task, there has been some work done regarding inter-modal correspondences between language and visual data \cite{DBLP:journals/corr/KarpathyF14} to align the two modalities. But in image captioning, the image and the text essentially have the same semantic information which is not the case for memes.

\section{Conclusion}
\label{conclusion}

In this paper, we explore the challenges in meme sentiment and humor analysis using deep learning. We compare the performance of large pretrained models with the simpler models that we train from scratch. Through our detailed analysis we show that the unconventional presence of both the modalities in memes reduces the effectiveness of domain adaptation or transfer learning techniques and discuss some ways to improve our models as future work.


\bibliographystyle{coling}
\bibliography{semeval2020}

\end{document}